\documentclass[preprint,12pt]{elsarticle}
\usepackage{amsmath}
\usepackage{graphicx}
\usepackage{adjustbox}
\usepackage{array,tabularx,booktabs,ragged2e}
\graphicspath{{figures/}}
\usepackage{pdfpages}
\usepackage[hidelinks]{hyperref}
\journal{World Patent Information}

\begin{document}
\begin{frontmatter}
\title{Graph-model gains, edge dependence, and the value of correct claim structure in patent litigation prediction}
\author[a]{Takao Arai\corref{cor1}}
\ead{af24l001@guh.u-hyogo.ac.jp}
\author[a,b]{Hiroyasu Inoue}
\cortext[cor1]{Corresponding author.}
\address[a]{Graduate School of Information Science, University of Hyogo, 7-1-28 Minatojima-minamimachi, Chuo-ku, Kobe, Hyogo 650-0047, Japan}
\address[b]{RIKEN Center for Computational Science, Kobe, Hyogo 650-0047, Japan}

\begin{abstract}
Structure-aware patent models have been reported to outperform models without the structure, with gains attributed to the encoded structure. We examine whether such gains can be attributed to the correct structure in one test case, claim dependency in the prediction of patent litigation involvement. The case suits this question because the legal role of dependency in patent disputes makes the correct references plausibly informative. Using 1,184,527 U.S. utility patents granted in 2004--2010 and five-year litigation involvement, we set two kinds of evidence from which the value of a structure is easily inferred against a direct test that retrained the same graph attention architecture with the actual references, randomized reference targets, or no edges. Confidence intervals included training randomness and family-clustered test-sample variation. Both kinds of evidence were present. Graph models had higher point estimates of the area under the receiver operating characteristic curve (AUC-ROC) than text-only baselines, by similar amounts with or without edges, and removing edges from trained models lowered AUC-ROC by 0.0055 (95\% CI 0.0017 to 0.0092). Retraining, however, showed no detectable advantage of the correct references: AUC-ROC differences were $+0.0004$ ($-0.0031$ to $+0.0039$) relative to randomized targets and $+0.0007$ ($-0.0035$ to $+0.0050$) relative to no edges. Point estimates of precision--recall performance and top-ranking lift favored the correct references, but all intervals included zero. Gains over text-only baselines and dependence on edges therefore did not establish the value of correct claim structure, although small effects cannot be excluded. Attributions to structure require retrained structural controls.
\end{abstract}

\begin{keyword}
Patent litigation \sep Patent claims \sep Claim dependency \sep Graph neural networks \sep Model evaluation \sep Patent analytics
\end{keyword}
\end{frontmatter}

\section{Introduction}
\label{sec:intro}

Predicting which patents will become involved in litigation supports litigation-risk management, portfolio valuation, and the screening of large patent collections. Earlier studies characterized litigated patents by their citations, claims, and owners~\cite{allison2004valuable,chien2011predicting}, and more recent work has applied machine learning to bibliographic and textual information~\cite{liu2018patent,juranek2024predicting}. Because claims define the scope of the right, claim text has become a central input to such models~\cite{liu2018patent,sakthivel2025multifeature}.

Patent analytics also uses how claims are structured. Studies have used the structure of claims, and of patent documents more broadly, as indicators of patent characteristics and outcomes: indicators of claim coverage built from independent and dependent claims and interpreted as offensive and defensive strength~\cite{liu2025new}, claim depth and the number of independent claims as predictors of licensing~\cite{alkhazaleh2024unveiling}, and structural features of patent documents, including the length of independent claims, as correlates of later information use~\cite{he2026patent}. In these studies, structure is summarized as patent-level variables and related to an outcome.

Predictive models have gone further by supplying the structure itself to the model. At the level of claims, Gao et al.~\cite{gao2024beyond} represented dependencies within and between claims as fine-grained graphs to predict the examination outcomes of individual claims, Tong et al.~\cite{tong2025what} weighted claims by their position in the dependency hierarchy and by their information density for patent similarity, and Yoo et al.~\cite{yoo2026phage} injected claim-dependency graphs into Transformer attention for patent classification, retrieval, and clustering. These studies reported improvements over variants without the structure and attributed them, at least in part, to the structure that the model encodes. Claim dependency has also been used in litigation-risk prediction: Sakthivel and Jose~\cite{sakthivel2025multifeature} used reference relations to group claims when aggregating a scope indicator, without separately evaluating the contribution of those relations, which was not the aim of their study. Other models represent relations between patents rather than between the claims of one patent, such as semantic-similarity graphs for early-stage valuation~\cite{liu2026beyond}, citation graphs for grant prediction~\cite{tong2026prediction}, and multi-relational graphs for identifying high-value patents~\cite{wang2025mrgce}; they concern a different structure and bear on this study only by analogy. An improvement of a structure-aware model as a whole is not the same as a contribution of the correct claim references, and which of the two a comparison shows depends on what it holds fixed.

For claim dependency, such attributions are plausible: dependency fixes the relative breadth of claims and provides fallback positions when broader claims are challenged (Section~\ref{sec:background-legal}). The references that define it, however, are written in the claim text, so representing them as graph edges can only make relationships already present in the text more accessible to a model. Whether doing so improves prediction is an empirical question that legal importance alone cannot settle.

Answering it requires separating the contribution of the correct structure from other differences between models. A graph model may outperform a text-only model because of components other than its edges, such as claim-wise encoding or the aggregation of claim representations. A trained model whose performance falls when its edges are removed at test time shows only that it responds to a change in its input, not that the edges improve prediction~\cite{hooker2019benchmark}. These difficulties are not specific to patents. Graph neural networks can rely on graphs that carry no task-relevant information~\cite{bechler2024graph}, attributing gains to structure requires comparison with structure-agnostic baselines~\cite{errica2020fair}, and in knowledge-graph-based recommendation, removing or randomly distorting the graph did not necessarily reduce accuracy~\cite{zhang2025kg4receval}. In patent analytics, recent work has likewise emphasized comparable and reproducible experiments and comparison with simple baselines~\cite{salampasis2025towards,bae2026patent}.

In the claim-structure studies whose full texts we examined, the comparisons did not separate the contribution of correct reference targets from that of adding structural information at all, or from accompanying changes in the model and its input. Gao et al.~\cite{gao2024beyond} compared fine-grained, claim-level, and dependency-free representations, but the dependency-free variant used a different classifier, and adding dependency between claims also added the text of referenced claims to the graph of a dependent claim. Tong et al.~\cite{tong2025what} compared claim representations weighted by hierarchical importance and information density with split claim sentences without these weights, without separating the two weights. Yoo et al.~\cite{yoo2026phage} compared their model with variants without the claim graph and with subsets of edge types or attention components. None of these studies compared the correct references with randomized ones under an otherwise identical model. Their comparisons support the performance of the proposed models on their tasks, which differ from the one studied here; they do not show how much of the improvement requires the references to be correct. Their task-specific gains are not reassessed here. Instead, we examine in a separate test case whether gains of the kind these studies report, together with a trained model's dependence on its edges, are sufficient to establish the predictive value of the correct references.

The study asks whether explicitly supplying the correct dependency references as a graph improves the prediction of patent-level litigation involvement beyond claim-wise text. Litigation involvement is a suitable test case because claim dependency plays a legal role in patent disputes (Section~\ref{sec:background-legal}), so that the correct references could plausibly carry information about this outcome. Using 1,184,527 U.S. utility patents granted in 2004--2010, chronological splits by grant year, and fixed post-grant observation windows, we retrain a graph attention model with the same architecture under the extracted references (TRUE), reference targets randomized within each patent (SHUFFLED), and no edges with readout of the independent claims only (NO-EDGE-IC). We also remove or randomize the edges of trained models at test time, and we estimate all three quantities on the same scale, with confidence intervals that combine variation across training seeds with variation in the test sample, clustered by patent family.

The study makes three contributions. First, retraining with the correct references yielded no detectable advantage in the area under the receiver operating characteristic curve (AUC-ROC), the primary metric, over randomized or absent references, and secondary metrics favored the correct references in point estimates only. Second, graph-model gains and edge dependence were nonetheless present, and the edge-free control performed close to the full model, so these kinds of evidence cannot by themselves establish the value of correct claim structure. Third, drawing on established evaluation principles, we summarize the checks needed before attributing predictive gains to claim structure, including elements specific to patent data such as family-clustered uncertainty and fixed observation windows. These results do not show that claim dependency lacks predictive value; they show that its added value could not be confirmed under the conditions studied here.

\section{Background}
\label{sec:background}

\subsection{Claim dependency}
\label{sec:background-legal}

Under 35 U.S.C. \S112, fourth paragraph, as in force before the Leahy--Smith America Invents Act and applicable to the patents studied here (now \S112(d)), a dependent claim refers back to a preceding claim, incorporates all of its limitations, and specifies a further limitation~\cite{usc112}. Dependency therefore fixes the relative breadth of claims: a dependent claim is never broader than the claim it references. Because each claim is presumed valid independently of the others~\cite{usc282}, a narrower dependent claim can survive when the claim it references is held invalid, so dependent claims serve as fallback positions in validity disputes~\cite{brean2019grading}. Differences between linked claims also inform claim construction~\cite{phillips2005}, and dependent claims are frequently among the claims found infringed~\cite{freilich2015pushing}. Empirically, litigated patents tend to have more independent claims and shorter independent claims than other patents~\cite{allison2004valuable,uspto2015patent}.

The references that define dependency are written in the claim text, typically as a phrase such as ``the method of claim 3.'' A model that encodes claims separately can register the phrase but not the content of the referenced claim; graph edges supply that link explicitly without adding information that is absent from the text.

\subsection{Three quantities}
\label{sec:quantities}

The quantity of interest in this study is the value of correct claim structure, defined below. Two other quantities are easily taken as evidence of it. The \emph{graph-model gain} is the difference in performance between a graph model and a text-only baseline, the kind of comparison reported for structure-aware models~\cite{gao2024beyond,tong2025what,yoo2026phage}; because the two models also differ in how claims are encoded, transformed, and aggregated, it cannot be attributed to the structure alone~\cite{errica2020fair}. \emph{Edge dependence} is the loss of performance when the edges of a trained model are removed or randomized at test time, as is commonly done to judge whether a model relies on part of its input~\cite{hooker2019benchmark}; it combines reliance on the information carried by the edges with the effect of inputs unlike those seen in training, and does not show that a model trained without the edges would perform worse. We measure both to see whether such evidence can appear when the value of correct structure is not detected. Edge dependence also checks that the trained models use their edges, so that an absence of value cannot be explained by models that ignore them.

The \emph{value of correct claim structure} is the difference between models with the same architecture that are retrained with the correct references and with control structures. We estimate it through two comparisons. Comparing the correct references with references whose targets are randomized within each patent, holding the number of references per claim fixed, addresses the correctness of the reference targets. Comparing the correct references with no edges, with the patent read out from its independent claims in both cases, addresses whether information from dependent claims, propagated along the correct references, improves on the independent claims alone. Neither control isolates correctness completely; Table~\ref{tab:conditions} lists what each preserves and changes.

\section{Data and methods}
\label{sec:methods}

\subsection{Patents, outcome, and timing of information}
\label{sec:data}

We analyzed U.S. utility patents granted in 2004--2010, extracted on 1 March 2026 from the PatentsView tables in the Google BigQuery public dataset~\cite{patentsview2026}. Because the release underlying this copy is not documented, we report the extraction date. We used grant dates, claim texts in claim-sequence order, current Cooperative Patent Classification (CPC) codes, and citations to U.S. patents. Of 1,340,620 patents granted in the period, we excluded design, plant, and reissue patents and identifiers that were not numeric, leaving 1,184,527 utility patents. The data were split by grant year into training (2004--2007; 639,428 patents), validation (2008--2009; 325,251), and test (2010; 219,848) sets.

Litigation involvement was taken from the March 2024 release of the USPTO Patent Litigation Docket Reports Data, which covers cases filed in U.S. federal district courts through 2020~\cite{toole2024technical}. A patent was labeled positive if at least one case was filed on or after its grant date and within five years (primary analysis) or ten years (sensitivity analysis); cases filed before grant were ignored, case types were not restricted, and a patent involved in several cases received a single label. Every window ends by 28 December 2020, so all windows are fully observed. Five-year positive rates were 0.67--0.78\% across the splits (Table~S1). The two analyses share the test patents, and five-year positives are also ten-year positives, so the ten-year analysis is a sensitivity analysis rather than a replication.

The predictors differ in when they become available. Claim texts and the number of backward citations are fixed at grant, whereas current CPC classifications were assigned after grant. Outcomes for training patents were observed over windows extending to 2012 (five-year) or 2017 (ten-year), beyond the grant dates of the test patents. The design therefore measures how well litigation involvement can be distinguished across chronologically ordered grant cohorts, not how well it could have been predicted at grant.

Patent families, used only for evaluation, were defined as connected components combining DOCDB simple families with continuation, continuation-in-part, divisional, and substitute relations recorded in PatentsView.

\subsection{Claim dependency graphs}
\label{sec:graphs}

Each patent was represented as a directed graph whose nodes are its claims and whose edges run from each dependent claim to the claim it references. Canceled claims were removed, and references were extracted with a regular expression that reads the claim numbers following ``claim'' or ``claims,'' separated by a comma, ``and,'' ``or,'' or a hyphen. Because a hyphen was treated as such a separator, a range such as ``claims 1-3'' yielded edges to its endpoints (claims 1 and 3) but not to the claims between them. A range written with ``to'' (``claims 1 to 3'') yielded an edge to its first number only, a number preceded by two separators (claim 3 in ``claims 1, 2, and 3'') was missed, and references that do not state a claim number (``any preceding claim'') were not resolved. Edges from claims whose references are written as ranges or without a claim number accounted for about 0.2\% of reference edges in PatentsView records. References to nonexistent claims and self-references were discarded, and edges forming cycles (269 in total) were removed, so all graphs were acyclic. Claims without outgoing edges were defined as independent claims, and the resulting independent-claim mask was held fixed in every condition. For patents granted in 2004, whose source texts lack claim numbers, numbers were assigned in order of appearance. Extraction accuracy was assessed against PatentsView dependency records for 2005--2010, by manual annotation of 20 patents (446 claims) selected by stratified sampling, and with a rule detecting claim-numbering shifts in 2004 patents (Supplementary Section~S1). Graph statistics are given in Table~S2.

\subsection{Features and models}
\label{sec:models}

Each claim was encoded separately with the frozen bert-for-patents model~\cite{srebrovic2020leveraging}, truncated to 512 tokens and averaged over the final-layer hidden states, giving a 1,024-dimensional vector. Patent-level features were the number of backward citations and indicators for the nine CPC sections.

The main model, CLAIMGAT, applies two graph attention layers~\cite{velickovic2018graph} (hidden size 256; four heads in the first layer and one in the second; self-loops and residual connections) to the claim vectors. Messages flow along the edges, from dependent claims to the claims they reference, so after two layers an independent claim can receive information from dependent claims up to two references away. The patent representation is an attention-weighted average over the independent claims only, which is concatenated with the patent-level features and passed to a two-layer perceptron.

To place CLAIMGAT among simpler models, we trained a model ladder: META (patent-level features only); BERT\_CONCAT (all claims concatenated and encoded as one text truncated to 512 tokens); BERT\_SPLIT (the mean of the claim vectors); CLAIMGCN (two graph convolutional layers~\cite{kipf2017semi} and the mean over independent claims); and CLAIMGAT-IC (graph attention layers and the mean over independent claims). Adjacent models differ in several components, so comparisons along the ladder are descriptive.

All models were trained with AdamW and binary cross-entropy weighted by the training positive rate, with early stopping and checkpoint selection on validation AUC-ROC. Each model was trained with five random seeds. In the ten-year analysis, TRUE, SHUFFLED, and NO-EDGE-IC were first trained with three seeds; two seeds were added after a difference in the area under the precision--recall curve (AUC-PR) had been observed. All runs of these models are reported. Hyperparameters and other implementation details are given in Supplementary Section~S2.

\subsection{Topology conditions}
\label{sec:conditions}

\begin{figure}[tbp]
\centering
\includegraphics[width=\linewidth]{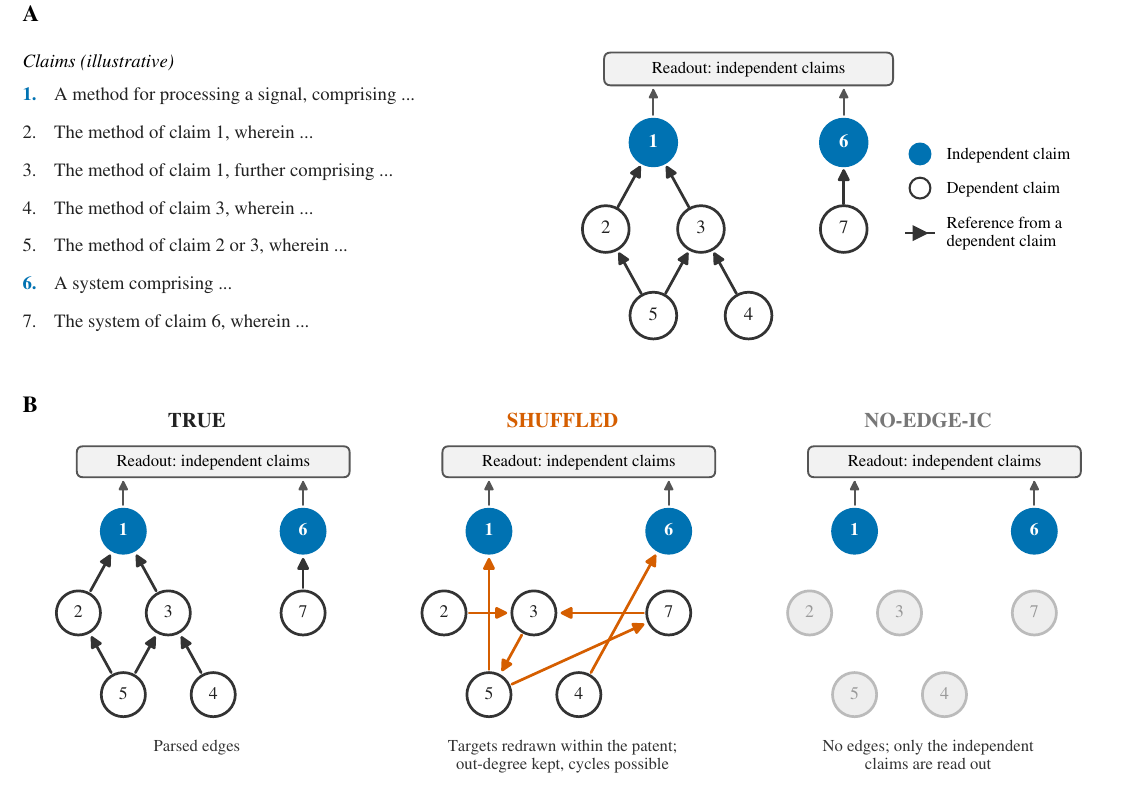}
\caption{Claim dependency graph and the three topology conditions. (A) An illustrative set of seven claims and its dependency graph; arrows point from dependent claims to the claims they reference, and information propagates in the same direction. Filled nodes are independent claims, whose representations alone are read out. (B) Edges supplied under TRUE (extracted references), SHUFFLED (targets redrawn within the patent; out-degree kept, cycles possible), and NO-EDGE-IC (no edges). The readout mask is common to all conditions.}
\label{fig:schematic}
\end{figure}

The central comparison retrains CLAIMGAT under three conditions that differ only in the edges supplied (Fig.~\ref{fig:schematic}; Table~\ref{tab:conditions}). TRUE uses the extracted references. In SHUFFLED, the source of each edge was kept and its target redrawn uniformly from the claims of the same patent, with self-loops redrawn up to eight times and then deleted; each run used one random graph per patent throughout training and evaluation, and different runs used different random graphs. NO-EDGE-IC supplies no edges, so that predictions rest on the independent claims and the patent-level features. NO-EDGE-IC was implemented by reversing the direction of the edges, which is equivalent under independent-claim readout because independent claims then receive no messages (Table~S7). In the test set, SHUFFLED preserved the out-degree of source claims in almost all patents but retained only 12.9\% of the original edges, changed what reaches the independent claims in 94\% of patents, and left 42\% of graphs acyclic (Table~S6).

TRUE versus SHUFFLED and TRUE versus NO-EDGE-IC have the interpretive roles described in Section~\ref{sec:quantities}, and Table~\ref{tab:conditions} lists what each control preserves and changes. SHUFFLED versus NO-EDGE-IC is reported descriptively. A supplementary condition, NO-EDGE-ALL, supplies no edges and reads out all claims (Supplementary Section~S2). Because it differs from TRUE in both the edges and the readout, it is not a primary control; it was trained with three seeds in the ten-year window, was not included in the analysis with combined intervals, and is reported only as descriptive values and in the earlier tests that treat patents as independent (Tables~S12, S17, and~S18).

\begin{table}[tbp]
\centering
\small
\caption{Properties preserved or changed under each topology condition.}
\label{tab:conditions}
\begin{adjustbox}{max width=\linewidth}
\begin{tabular}{>{\RaggedRight\arraybackslash}p{0.26\linewidth}>{\RaggedRight\arraybackslash}p{0.2\linewidth}>{\RaggedRight\arraybackslash}p{0.24\linewidth}>{\RaggedRight\arraybackslash}p{0.2\linewidth}}
\hline
Property & TRUE & SHUFFLED & NO-EDGE-IC \\
\hline
Edges & Extracted references & Targets redrawn within the patent & None \\
Out-degree of source claims & Original & Preserved (rare exceptions) & Zero \\
In-degree distribution and acyclicity & Original & Not preserved & Not applicable \\
Independent-claim readout & Fixed & Fixed & Fixed \\
Dependent-claim information at the readout & Along extracted references & Along random edges & None \\
\hline
\end{tabular}
\end{adjustbox}
\end{table}

\subsection{Evaluation and statistical inference}
\label{sec:evaluation}

AUC-ROC was the primary metric. Because fewer than 1\% of patents were positive, we also report, as secondary metrics, AUC-PR, computed as average precision~\cite{saito2015precision}, and lift among the top 0.5\%, 1\%, 5\%, and 10\% of the ranking.

Each of the three quantities was estimated as a difference in the expected performance of a single trained model (Table~\ref{tab:estimands}). For comparisons between retrained models, this is the difference between the condition means over the five training seeds; for edge dependence, it is the mean change in the five trained TRUE models when their edges are manipulated. The two pre-specified primary comparisons are TRUE versus SHUFFLED and TRUE versus NO-EDGE-IC in AUC-ROC.

\begin{table}[tbp]
\centering
\small
\caption{The three quantities, the comparisons that estimate them, and where they are reported.}
\label{tab:estimands}
\begin{adjustbox}{max width=\linewidth}
\begin{tabular}{>{\RaggedRight\arraybackslash}p{0.2\linewidth}>{\RaggedRight\arraybackslash}p{0.37\linewidth}>{\RaggedRight\arraybackslash}p{0.19\linewidth}>{\RaggedRight\arraybackslash}p{0.15\linewidth}}
\hline
Quantity & Comparison & Models & Results \\
\hline
Graph-model gain & TRUE and NO-EDGE-IC, each versus the text-only baselines BERT\_CONCAT and BERT\_SPLIT & Retrained; architectures differ & Section~\ref{sec:res-gain} \\
Edge dependence & Original edges versus edges removed, or reference targets randomized, at test time & Same trained TRUE models & Section~\ref{sec:res-dependence} \\
Value of correct claim structure & TRUE versus SHUFFLED and TRUE versus NO-EDGE-IC (primary comparisons) & Retrained; same architecture & Section~\ref{sec:res-value} \\
\hline
\end{tabular}
\end{adjustbox}
\end{table}

Inference rests on one 95\% confidence interval for each difference, which combines training randomness, through the variance of each seed mean, with test-sample variation. The latter is estimated by a bootstrap over patent families (2,000 resamples) from the covariance of the metric between different runs, so that test-sample error shared by the runs is counted once, as in studies in which several readers rate the same cases~\cite{obuchowski1995hypothesis}; in simulations with no true difference, the interval excluded zero in 4.2--5.4\% of replications (Supplementary Section~S3). Differences in AUC-PR and lift are ratios of geometric means across seeds minus one, with intervals on the log scale. Intervals were not adjusted for multiplicity; where several intervals bear on one statement, we report how many were examined. For each comparison we also report the difference detectable with 80\% power at a two-sided 5\% level.

Two further summaries are secondary. A seed-level exact permutation test (five runs against five; smallest attainable two-sided $p$, 0.008) asks whether a difference is consistent across training runs on this test set, ignoring test-sample variation. Differences between five-seed ensembles, with family-clustered bootstrap intervals, are conditional on the trained models. Ensembling raised AUC-ROC by different amounts in different conditions (in the five-year window, by 0.0015 for BERT\_CONCAT, 0.0069 for TRUE, 0.0083 for SHUFFLED, which uses a different random graph in each run, and 0.0090 for BERT\_SPLIT; Table~S17), so differences between ensembles combine the quality of a condition with the diversity of its runs, and we do not base conclusions on them.

The combined interval was adopted after earlier analyses had been examined. Those analyses had reported the two sources of uncertainty separately, through ensemble differences and seed-level tests with Holm adjustment~\cite{holm1979simple}, and had given no interval for graph-model gains or edge dependence; how intervals would be translated into wording was fixed before the combined interval was computed (Supplementary Section~S3). Earlier tests on ensemble predictions that treat patents as independent, including the DeLong test~\cite{delong1988comparing}, are retained in Table~S18 and are used in the Results only for the analysis by case type.

\subsection{Additional analyses}
\label{sec:additional}

Edge dependence was measured on the five trained TRUE models for the five-year window, with their parameters fixed, by removing all test-set edges or by randomizing reference targets as in SHUFFLED (three independent draws, averaged). Table~S11 also describes how the manipulations changed individual predictions. Because edge dependence and the value of correct structure are both measured from the TRUE models, their difference can be estimated directly: the edge-removal loss minus the difference between TRUE and NO-EDGE-IC equals the difference between NO-EDGE-IC and the TRUE models with their edges removed. We report this difference to compare the two quantities, not as a further quantity.

Two robustness analyses reused existing predictions without retraining. The first removed test patents sharing a family with a training or validation patent and repeated the estimates of the three quantities. The second restricted positives to patents involved in infringement actions brought by the patentee and, separately, examined positives involved only in other case types, using the case-type classification in the docket data~\cite{toole2024technical}.

Finally, after the test results had been examined, we compared the three topology conditions on the validation cohort, which had been used for checkpoint selection, using the same trained models. We regard this check as exploratory. Differences and intervals were computed as for the test cohort. We also report differences between five-seed ensembles with family-clustered intervals and seed-level permutation tests, without adjustment for multiple comparisons, and repeated the analysis after excluding validation patents that shared a family with a test patent.

\section{Results}
\label{sec:results}

\subsection{Graph-model gain}
\label{sec:res-gain}

Claim text accounted for most of the predictive performance. In the five-year window, mean AUC-ROC rose from 0.7395 for the model using only citation counts and CPC sections (META) to 0.8431 for the model that adds the concatenated claim text (BERT\_CONCAT; Table~\ref{tab:ladder}). This change is attributable to text, not to graph modeling.

The graph attention models had higher point estimates than both text-only baselines (Table~\ref{tab:ladder}; Fig.~\ref{fig:main}A). In the five-year window, expected single-run AUC-ROC was higher for TRUE than for BERT\_CONCAT by $+0.0044$ (95\% CI $-0.0009$ to $+0.0097$) and than for BERT\_SPLIT by $+0.0106$ ($+0.0015$ to $+0.0197$). The edge-free model showed gains of nearly the same size ($+0.0037$ and $+0.0098$ for NO-EDGE-IC), and the gains over BERT\_SPLIT, the only text-only baseline in the ten-year window, were of similar size there (Table~S8). Two of the six intervals excluded zero, both against BERT\_SPLIT.

The gain over BERT\_CONCAT was consistent across training runs: every TRUE run and every NO-EDGE-IC run scored higher than every BERT\_CONCAT run (seed-level $p = 0.008$ for both). Its interval nevertheless included zero because test-sample variation, not training randomness, accounted for about 90\% of the variance of the difference. The gain is therefore well supported for these models on this test set but is not established for a new sample of patents. In lift among the top 1\%, the graph models exceeded BERT\_CONCAT but not BERT\_SPLIT (Table~S9). Between five-seed ensembles in the five-year window, the gain of TRUE was about twice its single-run value over BERT\_CONCAT ($+0.0098$; family-clustered 95\% CI $+0.0044$ to $+0.0151$) but smaller over BERT\_SPLIT ($+0.0084$ against $+0.0106$), because ensembling helped the graph models more than BERT\_CONCAT and helped BERT\_SPLIT most (Section~\ref{sec:evaluation}; Table~S17).

\begin{table}[tbp]
\centering
\small
\caption{Predictive performance of the model ladder and the retrained topology conditions (mean $\pm$ SD over five training seeds).}
\label{tab:ladder}
\begin{adjustbox}{max width=\linewidth}
\begin{tabular}{lcccc}
\hline
 & \multicolumn{2}{c}{Five-year window} & \multicolumn{2}{c}{Ten-year window} \\
Model & AUC-ROC & AUC-PR & AUC-ROC & AUC-PR \\
\hline
\multicolumn{5}{l}{\textit{Model ladder}} \\
META & $0.7395 \pm 0.0012$ & $0.0268 \pm 0.0001$ & -- & -- \\
BERT\_CONCAT & $0.8431 \pm 0.0004$ & $0.0544 \pm 0.0023$ & -- & -- \\
BERT\_SPLIT & $0.8369 \pm 0.0071$ & $0.0562 \pm 0.0013$ & $0.8271 \pm 0.0026$ & $0.0670 \pm 0.0017$ \\
CLAIMGCN & $0.8404 \pm 0.0027$ & $0.0570 \pm 0.0041$ & $0.8284 \pm 0.0023$ & $0.0663 \pm 0.0040$ \\
CLAIMGAT-IC & $0.8451 \pm 0.0020$ & $0.0599 \pm 0.0013$ & $0.8327 \pm 0.0027$ & $0.0693 \pm 0.0030$ \\
CLAIMGAT (TRUE) & $0.8475 \pm 0.0015$ & $0.0605 \pm 0.0015$ & $0.8329 \pm 0.0047$ & $0.0716 \pm 0.0046$ \\
\multicolumn{5}{l}{\textit{Retrained topology controls}} \\
CLAIMGAT (SHUFFLED) & $0.8471 \pm 0.0026$ & $0.0590 \pm 0.0023$ & $0.8344 \pm 0.0014$ & $0.0674 \pm 0.0022$ \\
CLAIMGAT (NO-EDGE-IC) & $0.8468 \pm 0.0020$ & $0.0584 \pm 0.0026$ & $0.8352 \pm 0.0009$ & $0.0669 \pm 0.0025$ \\
\hline
\end{tabular}
\end{adjustbox}

\smallskip
\begin{minipage}{0.95\linewidth}
\footnotesize
Values are single-run means and SDs. --, not trained for the ten-year window. Adjacent models in the ladder differ in several design elements, so differences along the ladder are descriptive; ensemble values and earlier tests of adjacent steps are given in Tables~S17 and~S18.
\end{minipage}
\end{table}

\subsection{Edge dependence}
\label{sec:res-dependence}

\begin{figure}[tbp]
\centering
\includegraphics[width=\linewidth]{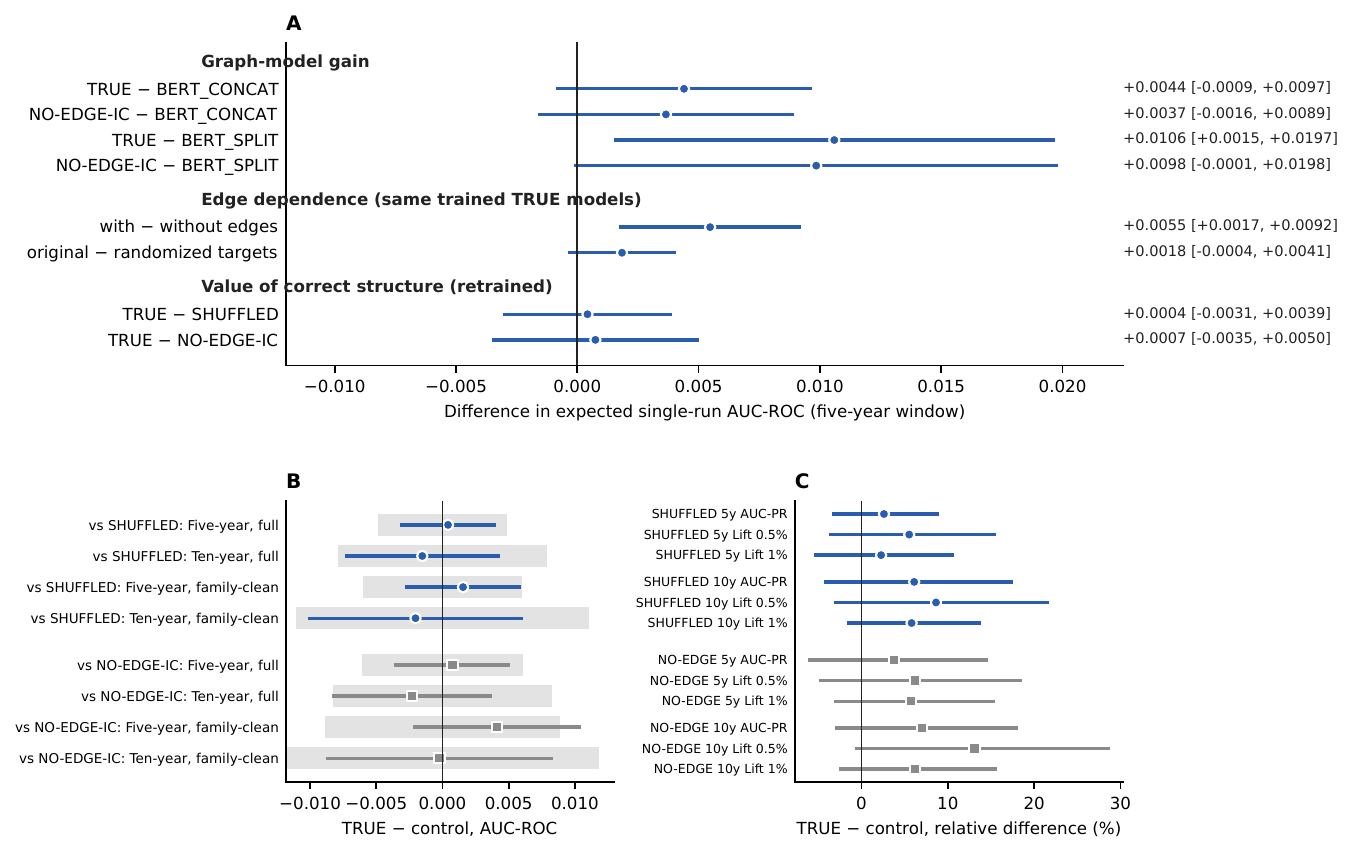}
\caption{The three quantities on a common scale. Points are differences in expected single-run performance and bars are 95\% confidence intervals that combine training randomness with test-sample variation across patent families; the vertical line marks zero. (A) AUC-ROC in the five-year window (full test set): graph-model gain over each text-only baseline, loss when the edges of the trained TRUE models were removed or their targets randomized at test time, and the retrained difference between TRUE and each control. (B) Value of correct claim structure (TRUE $-$ control) in AUC-ROC, in the five- and ten-year windows and on the family-clean test set; grey bands show $\pm$ the difference detectable with 80\% power. (C) Relative differences TRUE $-$ control in AUC-PR and in lift among the top 0.5\% and 1\% (full test set). In (B) and (C), circles denote comparisons with SHUFFLED and squares comparisons with NO-EDGE-IC.}
\label{fig:main}
\end{figure}

The trained TRUE models depended on their edges. With their parameters fixed, removing all edges from the five-year test data lowered expected single-run AUC-ROC by 0.0055 (95\% CI 0.0017 to 0.0092), and each of the five models lost performance (Fig.~\ref{fig:main}A; Tables~S8 and~S11). Randomizing the reference targets lowered AUC-ROC by 0.0018 ($-0.0004$ to $+0.0041$); the decrease occurred in every model, but its interval included zero. Edge removal also lowered AUC-PR and lift among the top 1\%, and both manipulations changed individual predictions and the composition of the top-ranked patents (Tables~S9 and~S11).

\subsection{Value of correct claim structure}
\label{sec:res-value}

Retraining with the correct references gave no detectable advantage in AUC-ROC (Fig.~\ref{fig:main}A and~B; Table~S8). In the five-year window, the difference in expected single-run AUC-ROC between TRUE and SHUFFLED, the comparison that bears most directly on the correctness of the references, was $+0.0004$ (95\% CI $-0.0031$ to $+0.0039$). Between TRUE and NO-EDGE-IC, which also concerns whether dependent-claim information reaches the readout, it was $+0.0007$ ($-0.0035$ to $+0.0050$). In the ten-year window, the differences were $-0.0015$ ($-0.0072$ to $+0.0042$) and $-0.0023$ ($-0.0082$ to $+0.0036$). Seed-level tests and differences between ensembles led to the same reading, and SHUFFLED and NO-EDGE-IC differed little from each other (Table~S8).

These results do not show that the correct references have no value. The differences that the design could detect with 80\% power were 0.0049 and 0.0061 in the five-year window and 0.0079 and 0.0082 in the ten-year window, and the upper confidence bounds remain compatible with advantages of up to about 0.004 to 0.005 in AUC-ROC.

Secondary metrics favored the correct references in point estimates, again without a detectable difference (Fig.~\ref{fig:main}C; Table~S9). In the twelve comparisons of TRUE with a control in AUC-PR and in lift among the top 0.5\% and 1\% (two controls, two windows), every point estimate favored TRUE, by 2.3\% to 13.1\%, and no interval excluded zero. The differences were larger in the ten-year window, for example in AUC-PR, $+6.1\%$ ($-4.1\%$ to $+17.4\%$) relative to SHUFFLED and $+7.0\%$ ($-2.9\%$ to $+17.9\%$) relative to NO-EDGE-IC. Summaries that omit one of the two sources of uncertainty were more favorable to TRUE for some ten-year comparisons; they are reported in Supplementary Section~S4 and Tables~S9 and~S18. The twelve estimates share one test set and correlated metrics and are not independent pieces of evidence. With detectable differences of 9\% to 20\%, a modest advantage among top-ranked patents would not have been detected. Among the top 5\% and 10\% of the ranking, mean lift was similar across the three conditions (Table~S10).

Because edge dependence and the value of correct structure are both measured from the TRUE models, the two quantities can be compared directly. The loss from edge removal exceeded the retrained difference between TRUE and NO-EDGE-IC by 0.0047 (95\% CI 0.0012 to 0.0083; Table~S8): a model trained without edges recovered most of what the trained models lost when their edges were removed. This difference is arithmetically the first quantity minus the second and adds no evidence beyond them. The corresponding difference for randomization was 0.0014 ($-0.0018$ to $+0.0046$).

\subsection{Robustness}
\label{sec:res-family}

\paragraph{Overlap of patent families}
Because the data were split by grant year rather than by family, 37,557 (17.1\%) of the test patents, including 667 (45.1\%) of those with five-year litigation involvement, shared a family with a training or validation patent. Excluding them (182,291 patents; 813 positives) lowered mean AUC-ROC by 0.0153 to 0.0194 for the models that use claims and by 0.0443 for META (Table~S12). On this family-clean set (Table~S13), the difference between TRUE and SHUFFLED remained close to zero ($+0.0015$; 95\% CI $-0.0027$ to $+0.0058$), whereas that between TRUE and NO-EDGE-IC was larger than on the full test set ($+0.0041$, $-0.0021$ to $+0.0103$, against $+0.0007$; unadjusted seed-level $p = 0.040$). Correspondingly, the gain over BERT\_CONCAT was larger in point estimates for TRUE ($+0.0085$; $+0.0015$ to $+0.0154$) than for NO-EDGE-IC ($+0.0044$), and the loss from edge removal no longer clearly exceeded the retrained difference between TRUE and NO-EDGE-IC (difference, 0.0021; $-0.0027$ to $+0.0070$). These estimates are less precise than those for the full test set (detectable differences of 0.0060 and 0.0089 for the two primary comparisons), but they suggest that the comparison with the edge-free control, unlike that with randomized references, may depend on whether patents related to earlier patents are included.

\paragraph{Case type and graph construction}
\label{sec:res-robust}
Restricting positives to patents involved in infringement actions brought by patentees (removing 137 and 195 positives in the five- and ten-year windows) left the comparisons of TRUE with the controls essentially unchanged (Table~S14; ensemble predictions, with intervals that treat patents as independent). For positives involved only in other case types, the five-year difference relative to SHUFFLED was $-0.0085$ ($-0.0167$ to $-0.0004$; Holm-adjusted $p = 0.082$), but these positives were too few to examine case types adequately. The extracted dependency relations agreed closely with PatentsView records for patents granted in 2005--2010 (edge-level F1, 0.998); errors found in manual annotation were confined to patents granted in 2004 with claim-numbering shifts, which accounted for 0.83\% of the dependency edges in the training data (Supplementary Section~S1).

\subsection{Exploratory check on the validation cohort}
\label{sec:res-valid}

After the test results had been examined, we compared the topology conditions on the validation cohort (325,251 patents granted in 2008--2009; 2,281 and 3,127 positives in the five- and ten-year windows), using the same trained models. Because this cohort had been used for checkpoint selection, the check is exploratory.

Point estimates resembled those in the test cohort. Relative differences in AUC-PR and in lift among the top 0.5\% and 1\% favored TRUE in all twelve comparisons; with intervals computed as for the test cohort, one excluded zero (ten-year AUC-PR relative to SHUFFLED, $+7.8\%$; 95\% CI $+1.8\%$ to $+14.2\%$; Table~S15). AUC-ROC differences were $+0.0000$ and $+0.0013$ in the five-year window and $+0.0014$ and $+0.0020$ in the ten-year window, all with intervals including zero; the ten-year differences, negative in the test cohort, were thus slightly positive here.

Differences between five-seed ensembles, which are conditional on the trained models and are not used as evidence (Section~\ref{sec:evaluation}), excluded zero in 9 of 24 family-clustered intervals (six metrics, two controls, two windows), all in favor of TRUE and including AUC-ROC (Table~S16). Results after excluding validation patents that share a family with a test patent are given in Supplementary Section~S6.

\section{Discussion}
\label{sec:discussion}

\subsection{Distinguishing three quantities}
\label{sec:disc-three}

The three quantities distinguished in Section~\ref{sec:quantities} led to different answers (Fig.~\ref{fig:main}A). The graph attention models had higher point estimates than the text-only baselines (Section~\ref{sec:res-gain}), and the trained models lost performance when their edges were removed (Section~\ref{sec:res-dependence}). Taken alone, these two results would suggest that the dependency structure contributes to prediction. Retraining the same architecture with correct, randomized, or absent references, however, showed no detectable advantage of the correct references in AUC-ROC, the primary metric (Section~\ref{sec:res-value}).

The comparisons answer different questions. The edge-free control uses the same architecture as the full model but receives no edges, so that information from dependent claims does not reach its readout. It nonetheless performed close to the full model and exceeded the text-only baselines by similar amounts. Comparable point-estimate gains over the text-only baselines were thus observed without the dependency edges, and without the dependent-claim information that the edges carry to the readout, although an additional contribution from the correct edges cannot be excluded. Which of the remaining differences between the graph models and the text-only baselines accounts for the gain, such as claim-wise transformation, attention-based aggregation, or restriction of the readout to independent claims, cannot be determined from these comparisons, which were not designed to isolate them. Edge dependence, in turn, shows that a model trained with edges uses them, not that a model trained without them would perform worse (Section~\ref{sec:quantities}). In our data the two measurements differed: the loss from edge removal exceeded the retrained difference between TRUE and NO-EDGE-IC (Section~\ref{sec:res-value}), although not clearly so on the family-clean test set (Section~\ref{sec:res-family}). Only the retrained comparison asks whether supplying the correct references improves the predictions that a model can learn, and this comparison did not show a detectable advantage.

Secondary metrics qualify this picture without changing it. Point estimates of AUC-PR and top-ranking lift favored the correct references, most visibly in the ten-year window, but no interval excluded zero; the estimates are compatible with a modest advantage among top-ranked patents that this design could not detect. On the family-clean test set, the difference from the edge-free control increased, whereas that from randomized references remained close to zero, so that result bears on whether information from dependent claims reaches the readout rather than on the correctness of the references (Section~\ref{sec:res-family}). Summaries that omit training randomness or test-sample variation, which were more favorable to TRUE in some ten-year comparisons, are not used as evidence (Section~\ref{sec:evaluation}).

The exploratory check on the validation cohort gave a similar picture (Section~\ref{sec:res-valid}): point estimates favored the correct references, but only one of the twelve intervals for AUC-PR and top-ranking lift, and none for AUC-ROC, excluded zero. These results are compatible with a small advantage of the correct references but do not establish it. The validation cohort had been used for checkpoint selection, the same trained models were scored on both cohorts, the comparisons were made after the test results had been examined, and the intervals were not adjusted for multiple comparisons. A small advantage of the correct references therefore remains a hypothesis, to be tested with new training runs on an independent cohort.

\subsection{Interpreting the absence of a clear advantage}
\label{sec:disc-why}

Several explanations are consistent with the results, and they are not mutually exclusive. First, the legal significance of claim dependency within patent disputes may not translate into predictive value for litigation involvement: dependency bears on claim scope, validity, and interpretation once a dispute arises, whereas whether a patent becomes involved at all also reflects the enforcement decisions and economic interests of the parties.

Second, much of the relevant claim information may already reside in the independent claims. The edge-free control, which reads out only those claims, performed close to the full model; measures based on independent claims have served as indicators of patent scope~\cite{marco2019patent,kuhn2019how}; and Allison et al.~\cite{allison2004valuable} found the number of independent claims, but not of dependent claims, to be associated with litigation. None of these observations is decisive, because the edge-free control also receives the patent-level features and Allison et al.\ counted claims rather than modeling their text.

Third, claim-wise text representations may already make accessible much of what explicit encoding would add: the text of a dependent claim states the claim it references, its subject matter, and its additional limitations, and lacks only the content of the referenced claim. Designs in which a dependent claim inherits the text of the claims it references, as in Gao et al.~\cite{gao2024beyond}, change the information available for each claim, and our results do not speak to them or to claim-level outcomes.

Fourth, the model may not exploit the structure fully: claims were encoded separately by a frozen encoder, and with two attention layers, information more than two references away from an independent claim could not reach the readout. Finally, the power to detect small differences was limited (Section~\ref{sec:disc-limitations}). Because we did not measure the overlap between dependency information and the text representations, and evaluated one family of models, the analysis cannot separate limitations of the model from weakness of the information itself.

\subsection{Implications for patent analytics}
\label{sec:disc-implications}

For studies that add claim structure, or other structure, to patent prediction models, the results indicate that improvements over text-only baselines and sensitivity to edge removal should not be read, on their own, as evidence that the correct structure carries predictive value. Table~\ref{tab:checklist} summarizes checks that follow from established principles for evaluating graph models and machine-learning benchmarks~\cite{errica2020fair,bechler2024graph,hooker2019benchmark,bouthillier2021accounting,zhang2025kg4receval}, together with their adaptation to patent data. The choice of estimand also mattered in our data: because averaging seeds helped the conditions by different amounts, the gain between ensembles was larger than the single-run gain over one text-only baseline and smaller over the other (Section~\ref{sec:res-gain}), so a gain reported for ensembles should not be read as the gain of a single model. None of these checks is new in itself. The contribution here is to show, for a legally meaningful structure in patent data, that applying them changes the conclusion that the more common comparisons would suggest.

These checks concern predictive claims. Indicators of claim structure, such as those of Liu and Chang~\cite{liu2025new} and Alkhazaleh and Mykoniatis~\cite{alkhazaleh2024unveiling}, can describe drafting strategy or scope without adding predictive information beyond claim text. Whether they add such information for a given outcome is the question that retrained structural controls are designed to answer.

\begin{table*}[p]
\centering
\footnotesize
\caption{Checks before attributing predictive gains to claim structure in patent analytics.}
\label{tab:checklist}
\setlength{\tabcolsep}{4pt}
\begin{tabularx}{\linewidth}{@{}>{\RaggedRight\arraybackslash}p{0.21\linewidth}>{\RaggedRight\arraybackslash}p{0.38\linewidth}>{\RaggedRight\arraybackslash}X@{}}
\toprule
Check & Rationale and sources & Adaptation to patent data (this study) \\
\midrule
Compare with retrained structural controls of the same architecture & Gains over structure-free baselines can arise from model components; models can use uninformative graphs~\cite{errica2020fair,bechler2024graph,zhang2025kg4receval} & Reference targets randomized within each patent with out-degree preserved; edge-free control with the same readout \\
\addlinespace[2pt]
Report what each control preserves and changes & Randomization alters more than correctness & SHUFFLED preserves out-degree but not in-degree or acyclicity; NO-EDGE-IC also removes dependent-claim information from the readout \\
\addlinespace[2pt]
Distinguish test-time ablation from retrained comparison & Removing inputs at test time confounds reliance with distribution shift~\cite{hooker2019benchmark} & Edge removal used only as a measure of edge dependence \\
\addlinespace[2pt]
Quantify uncertainty from both the test sample and training randomness, for expected single-run performance & Training randomness can be comparable to differences between methods~\cite{bouthillier2021accounting}; ensembles benefit conditions unequally & Intervals combining seed variance and a family-clustered bootstrap; five training seeds per condition; differences between ensembles reported separately \\
\addlinespace[2pt]
Use temporal splits and fixed observation windows & Random splits and variable follow-up can inflate or bias performance & Splits by grant year; litigation counted within fixed windows after grant; family overlap across splits examined \\
\addlinespace[2pt]
Pre-specify the primary metric & Rare outcomes make metrics that emphasize the top of the ranking sensitive and variable~\cite{saito2015precision} & AUC-ROC primary; AUC-PR, lift, and the validation-cohort check treated as secondary or exploratory \\
\addlinespace[2pt]
Validate structure extraction & Parser errors change the structure being tested & Agreement with PatentsView and manual annotation \\
\bottomrule
\end{tabularx}
\end{table*}

\subsection{Limitations}
\label{sec:disc-limitations}

Statistical uncertainty remains. With five training runs per condition, differences in AUC-ROC below about 0.005 to 0.006 (five-year window) or 0.008 (ten-year window), and relative differences in AUC-PR and top-ranking lift below about 9\% to 20\%, could not be distinguished from zero. The intervals describe variation across patent families and training seeds, not across grant cohorts, are not adjusted for multiplicity, and rest on a variance across seeds that five runs estimate imprecisely. The absence of a clear difference is therefore neither evidence of equivalence nor evidence that explicitly encoding the correct references cannot improve prediction. Neither control isolates correctness completely (Table~\ref{tab:conditions}).

The systematic evaluation was limited to CLAIMGAT-based models, which encode claims separately with a frozen encoder; representations that incorporate the referenced claims' text, finer-grained graphs, and other architectures were not evaluated under the same controlled design. Whether the correct references add predictive value in the tasks for which structure-aware claim models have been proposed, such as examination outcomes, patent similarity, and classification, was not examined; the comparisons used here can be applied to them. Patent-level features were limited to the number of backward citations and current CPC sections; patentee attributes and post-grant events were not included, and their inclusion can reduce the additional contribution of text-based information~\cite{juranek2024predicting}. The included features are unlikely to carry much of what dependency references encode. Because current CPC sections were assigned after grant and outcomes for training patents were observed beyond the grant dates of the test patents, the analysis estimates discrimination across chronologically ordered grant cohorts rather than prediction at grant (Section~\ref{sec:data}); as all conditions share these features and labels, this bears mainly on absolute performance. The outcome is involvement in U.S. federal district court litigation and should not be read as a measure of patent quality or value. The test cohort comprises patents granted in a single year, the graphs may contain parser errors, and some analyses were added after initial results had been examined: two training seeds in the ten-year analysis, the family-clustered bootstrap, the validation-cohort check, and the combined intervals. The conclusions therefore rest on effect sizes, intervals, and the consistency of several analyses rather than on any single test.

\section{Conclusions}
\label{sec:conclusions}

Claim dependency is a legally meaningful structure, and its references are written in the claim text. We asked whether supplying these references explicitly as the correct graph improves the prediction of patent-level litigation involvement, as a test of whether the gains of structure-aware models can be attributed to the correct structure. For U.S. utility patents granted in 2004--2010, the graph models had higher point estimates than text-only baselines, by similar amounts with or without edges, and the trained models lost performance when their edges were removed. The value of correct claim structure, measured by retraining the same architecture with randomized reference targets or no edges, was nevertheless not distinguishable from zero in AUC-ROC ($+0.0004$ and $+0.0007$; upper confidence bounds $+0.0039$ and $+0.0050$). Point estimates of AUC-PR and of lift among the top 0.5\% and 1\% of patents favored the correct references, but every interval included zero; a small advantage of this kind remains possible but unestablished.

Graph-model gains and edge dependence therefore did not establish the value of correct claim structure. This conclusion is limited to the text representation, model, features, and outcome studied here, and it does not imply that claim dependency lacks predictive relevance. Attributing predictive gains to claim structure requires comparison with retrained controls of the same architecture and an assessment of uncertainty from both the test sample and training randomness.

\section*{Funding}
This work was supported by the Japan Society for the Promotion of Science (JSPS) KAKENHI [grant numbers JP25K01454, JP26K22092, JP26K00349, JP23K25520, JP24K00247, JP24K01108]. The funder had no role in study design, data collection and analysis, decision to publish, or preparation of the manuscript.

\section*{Declaration of competing interest}
The authors declare that they have no known competing financial interests or personal relationships that could have appeared to influence the work reported in this paper.

\section*{Data availability}
Patent and litigation data are publicly available from the sources cited in Section~\ref{sec:data}. The code for data extraction, model training, and analysis will be made publicly available.

\section*{Declaration of generative AI and AI-assisted technologies in the manuscript preparation process}
During the preparation of this work the authors used Claude (Anthropic) in order to draft the English text of the manuscript from an outline prepared by the authors, to discuss the analysis plan and the statistical methods, to obtain critical comments on the manuscript and the analysis plan, to search for additional literature and compile bibliographic information, and to develop analysis code. After using this tool, the authors reviewed and edited the content as needed and take full responsibility for the content of the published article, including the analyses, the reported results, and their interpretation.

\bibliographystyle{elsarticle-num}
\bibliography{references_draft}

\clearpage
\includepdf[pages=-]{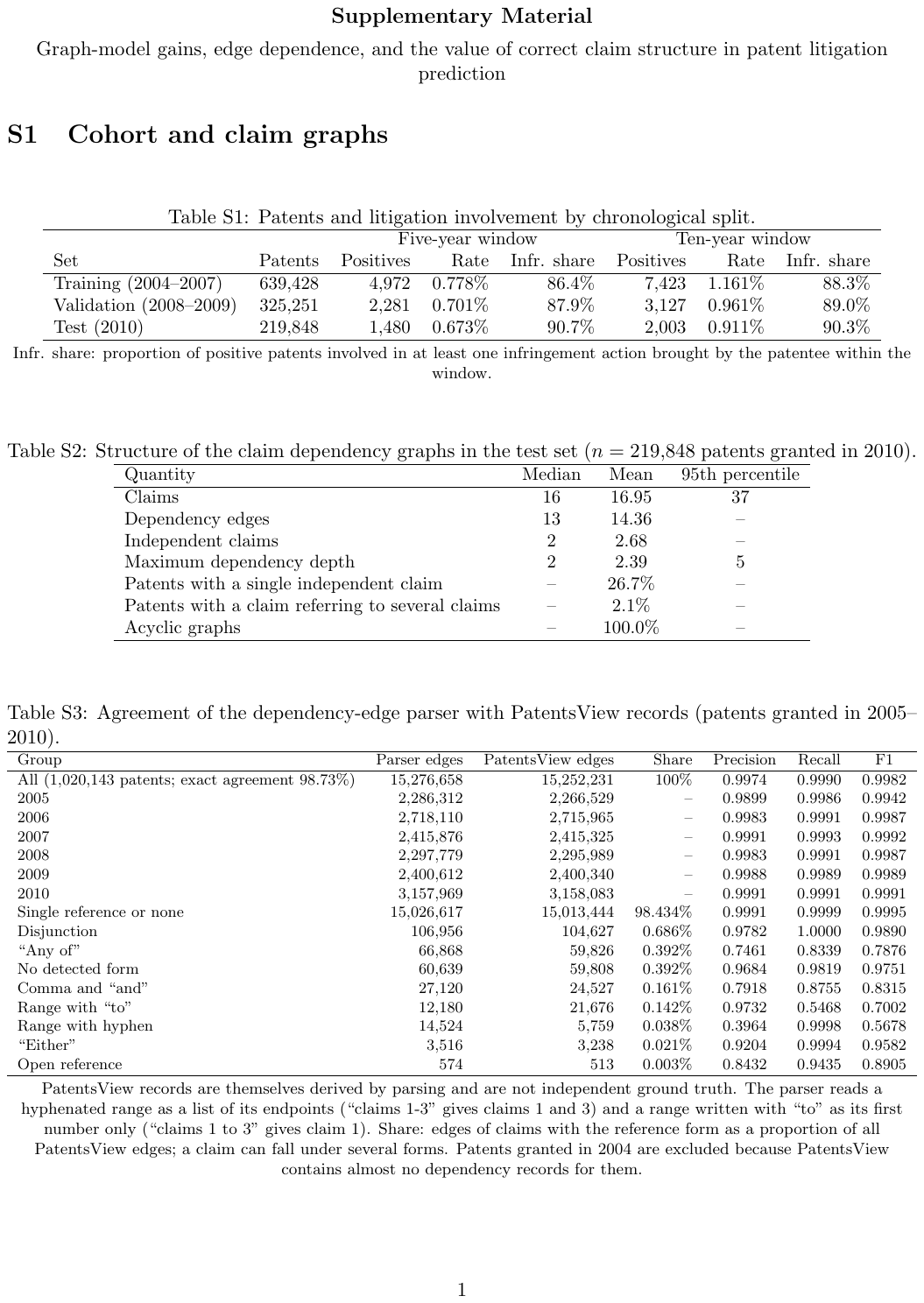}
\end{document}